%% file: neurips_distshift_2023.tex
\documentclass{article}

    \PassOptionsToPackage{numbers, compress}{natbib}

\usepackage[final]{neurips_distshift_2023}

\usepackage[utf8]{inputenc} % allow utf-8 input
\usepackage[T1]{fontenc}    % use 8-bit T1 fonts
\usepackage{hyperref}       % hyperlinks
\usepackage{url}            % simple URL typesetting
\usepackage{booktabs}       % professional-quality tables
\usepackage{amsfonts}       % blackboard math symbols
\usepackage{nicefrac}       % compact symbols for 1/2, etc.
\usepackage{microtype}      % microtypography
\usepackage{xcolor}         % colors
\usepackage{booktabs}
\usepackage{amsthm, amsmath, amssymb}
\usepackage{mathtools}
\usepackage{graphicx,subcaption,float}
\usepackage{breqn}
\usepackage[autostyle]{csquotes}
\usepackage[british]{babel}
\usepackage{breakcites}
\usepackage{bbm}
\usepackage{multirow}
\usepackage{hyperref}
\usepackage{comment}
\input{shortcuts.tex}
\usepackage{natbib}

\title{Tackling Concept Shift in Text Classification using Entailment-style modeling}

\author{%
  Sumegh Roychowdhury\thanks{Equal Contribution.} \\
  \texttt{sumegr@amazon.com} \\
  \And
   Karan Gupta\footnotemark[1] \\
  \texttt{karaniis@amazon.com} \\
  \And
  Siva Rajesh Kasa\footnotemark[1] \\
  \texttt{kasasiva@amazon.com} \\
  \And
  Prasanna Srinivasa Murthy \\
  \texttt{sprsn@amazon.com} \\
  \And
  Alok Chandra \\
  \texttt{alokchan@amazon.com} \\
}

\begin{document}

\maketitle

\begin{abstract}
  Pre-trained language models (PLMs) have seen tremendous success in text classification (TC) problems in the context of Natural Language Processing (NLP). In many real-world text classification tasks, the class definitions being learned do not remain constant but rather change with time - this is known as \textbf{ concept shift}. Most techniques for handling concept shift rely on retraining the old classifiers with the newly labelled data. However, given the amount of training data required to fine-tune large DL models for the new concepts, the associated labelling costs can be prohibitively expensive and time consuming. In this work, we propose a reformulation, converting vanilla classification into an entailment-style problem that requires significantly less data to re-train the text classifier to adapt to new concepts. We demonstrate the effectiveness of our proposed method on both real world \& synthetic datasets achieving absolute F1 gains upto \textbf{$\sim$7\%} and \textbf{$\sim$40\%} respectively in few-shot settings. Further, upon deployment, our solution also helped save \textbf{75\%} of labeling costs overall.
\end{abstract}

\section{Introduction}
\label{intro}
% Text Classification (TC) is the process of segregating textual units (such as queries, documents, social media texts, etc.) into pre-defined labels/classes. Text classifier models are usually trained through a supervised learning approach, wherein labeled textual units are fed into the models during training stage, and the trained classifier is later used to predict the labels of unseen datapoints in the inference stage.  TC is a widely studied area in Natural Language Processing (NLP) with applications in Sentiment Analysis (SA), Topic
% Labeling (TL), News Classification (NC), etc \cite{li2022survey}. 
%  Deep learning models have shown tremendous impact in text classification problems, specifically with the advent of large-scale Transformer-based Pre-trained Language Models (PLMs) \cite{minaee2021deep,vaswani2017attention}. PLMs are first pre-trained on much
% larger amounts of text corpora and are later fine-tuned using task-specific data. 
% %% To Do: rephrase the above sentence
% Autoencoding PLMs such as BERT \cite{DBLP:conf/naacl/DevlinCLT19} and its variants have improved the state-of-the-art in many
% downstream NLP tasks, including TC. Further, PLMs with large model size have been shown to have few-shot learning capabilities i.e. with only a few task-specific labeled data they can perform fairly well \cite{wang}. 
 
%  % BERT is one of the widely autoencoding PLMs used in text classification. There have been several variants of BERT depending on training data, model size, domain knowledge, etc.  

Concept shift (aka label shift) occurs in a data stream when 
% the distribution of classes for a given input textual unit changes over time. 
% Concept shift occurs when 
the contextual relationship between textual inputs (X) and labels (Y) changes with time, resulting in a change in the conditional probability of Y given X. Concept shift has been been studied extensively in classical machine learning problems \cite{klinkenberg2002concept,klinkenberg2004learning,kolter2007dynamic,lindstrom2010handling}. Concept shift in text classification can occur due to various reasons such as  
sudden change in external circumstances (e.g. occurrence of pandemics \cite{muller2020addressing}), gradual change in semantics (e.g. words taking on newer meanings \cite{bjerva2020back}), etc. 
% % Examples of changing concepts can be seen
% in a variety of real-world applications: , customer buying
% preferences can be influenced by fashion trends or seasonal
% inclinations, and changing users’ interests can impact
% on information filtering. 
Consider the case of classifying a healthcare research article as relevant/irrelevant to a user. Under the usual circumstances, the user may not be interested in healthcare research but during pandemics, his/her preferences could change. Depending on the direction and extent of concept shift, re-training of the classifier models is essential to match the changing concepts in the stream. If concept shift is not kept track of, it can lead to malfunction of downstream systems, loss of revenue, erosion of trust in the classifier systems, etc.

In the context of text classification (TC) using PLMs, tackling concept shift requires relabelling of the datapoints under the new concepts. While PLMs demonstrate few-shot learning capabilities, the amount of labelled data required primarily depends on the difficulty of the new concepts to be learnt and on the size of the PLM. As mentioned before, obtaining new labelled data is often laborious, time consuming and expensive in real-world settings. In this paper, we propose an Entailment-style approach \cite{roy2023hate} to overcome this data requirement. Further, there is a widespread consensus on the lack of real-world benchmark datasets for TC task to evaluate the various methods proposed for tackling concept-shift \cite{lindstrom2010handling,narasimhamurthy2007framework}, and hence, in addition to synthetic datasets, we also curate a real world dataset to study concept shift and benchmark our model performances on the same. To the best of our knowledge, no prior work has explored tackling sudden concept shift in the context of textual classification data. Refer to Appendix~\ref{related_work} for more details.

Our contributions are as follows:
% a) We introduce \textbf{\textit{RetailQueries Shift dataset}} for concept shift research in the text-classification domain. To the best of our knowledge, this is the first real-world multilingual dataset for NLP-based text classification concept-shift benchmark; it contains 200k manually annotated (keyword, product) relevance datapoints from an e-commerce website, 
(a) We show that \textbf{\textit{reformulating vanilla classification as an entailment-style task}} can help augment PLMs performance in tackling concept shift (b) We provide ablations to highlight PLMs  \textbf{\textit{few-shot capabilities for tackling concept shift}} in text classification. Our proposed approach leads to significantly better performance vis-a-vis a vanilla finetuning of PLMs, under the new concepts, as demonstrated on both real \& synthetic datasets. We defer further details motivating our work to Appendix~\ref{related_work} due to space constraints.
\input{anecdotes}
\section{Dataset Details}
\label{datasets}

\textbf{RetailQueries Shift} - For creating this data\footnote{Since its proprietary data, we cannot make it public without proper licensing. We will work on releasing the datasets in future iterations.}, we randomly sample 200k query-product pairs across 12 languages from an e-commerce store retail catalog. For each query, there are a number of products retrieved. Each of these (query, product) pairs are classified into 4 relevance categories: \textbf{exact (E)} - product fully satisfies the query intent, \textbf{substitute (S)} - not exact but functionally equivalent, \textbf{complement (C)} - not intended but could be useful and \textbf{irrelevant (I)} - does not satisfy query intent. For example, given query ``iphone'' the product ``iphone 11'' would be exact, ``samsung galaxy'' would be substitute, ``airpods pro'' would be complement and ``lord of the rings'' would be irrelevant. 

Now we describe the type of concept shift that occurs in the dataset. Annotators receive crisp guidelines as to what relevance labels (E/S/C/I) should be assigned to query-product pairs. Based on observed customer behavior, a change in the labeling guidelines was introduced, as to what is to be considered ``irrelevant" to better match the customer intent. So \textbf{pre-shift} all labels in our 200k dataset were \textbf{``irrelevant''}. Some labels changed post-shift due to guideline change while some remained same. The major changes were - \textbf{(a)} the product type intended by the query must be matching with the retrieved product for it to be relevant (E/S/C) under old guidelines. This condition was relaxed under new guidelines. Now, if a product can provide the same utility the customer is looking for, the query-product pair can be classified as exact. \textbf{(b)} Under new guideline, certain constraints on product attributes are relaxed. For e.g. attributes viz. Size, Age, etc. are relaxed and products at one level higher or lower will be considered substitute, while under old guidelines, they would be considered irrelevant. \textbf{(c)} under new guidelines, if brand intent matches then the product can be marked complement even if it has a different purpose than what's intended in query. 

\textbf{AGNews Shift} - We use the AGNews Topic Classification \cite{agnews} dataset where each news article is divided into 4 topics - World, Business, Sports and Sci-Tech and synthetically induce concept shift here in the following manner - we say that before the shift news related to topics World and Business are \textit{relevant} to the user while Sports and Sci-Tech are \textit{irrelevant}. After the shift, we switch the labels in a way that Sports and Sci-Tech become \textit{relevant} while World news becomes \textit{irrelevant}. Business news we keep the same label (\textit{relevant}) to induce a bit more complexity in the artificial shift.

For further details regarding dataset statistics refer to Table~\ref{table1} and Appendix ~\ref{app:data}.

\section{Proposed Approach}
\label{proposed_approach}
\input{main_figure}

Let the $\cl = \sqrbrkt{L_1,\dots,L_K}$ be the set of labels. The training dataset can be segregated based on the labels as $\cd_{tr} = \sqrbrkt{D_1,D_2,\dots,D_{K} }$ where $\left\{x_{i}^k \right\}_{i=1}^{n_k}$ is the set of training data available for the $k$-th label $L_k$. Let the concept shifted data be $\td{\cd}_{tr} = \sqrbrkt{\td{D}_1,\td{D}_2,\dots,\td{D}_{K} }$ where $\left\{ {\td{x}_{i}}^k\right\}_{i=1}^{n_k'}$ is the set of training data available for the $k$-th label $L_k$ after the concept shift. 
% Let $\cd'_{tst}$ be the corresponding test data after concept shift. 
Let $\cm$ be the finetuned model on the data $\cd_{tr}$.
For the concept-shifted data $\cd'_{tr}$, during training we do the following entailment-style data augmentation:

For a given datapoint $x$ whose pre-shift label is $L_j$ and post-shift label is $L_{j'}$, we create a total of $K$ augmented samples ($s_{k}$) with binary labels as follows:

\begin{equation}
    s_{k} = \{prompt(L_{k}) + x,\ \mathbbm{1}_{j'}(k) \} \ \forall \ k \in \sqrbrkt{1,2,\dots,K}
\end{equation}
where $\mathbbm{1}_{j'}()$ is the indicator function which
takes the value one only when the argument is $j'$, otherwise it is zero. `+' here refers to concat() operation (check \S~\ref{sec:impl}). Also $prompt()$ is defined as :

\begin{align}
    prompt(L_k) = \label{eq:prompting}
    \begin{cases}
        remained \ \{L_k\} match  & , L_k = L_j \\
        changed \ to \ \{L_k\} match  &  , o.w. 
    \end{cases}
\end{align}
Essentially, we create $(K-1)$ negative samples and 1 positive sample for each datapoint. Finally, the problem reduces to the question - Does $x$ entail $prompt(L_k)$ or not ? 

After creating these $K \times \sum_{k=1}^{K} {n_k}$ augmented examples, we finetune $\cm_0$ on a binary classification task. Let $\td{\cm}$ be the finetuned model.
During inference, the predicted label for an datapoint $\td{x}_i$ is obtained by taking an argmax over the probabilities output by the binary classifier $\td{\cm}$ over the augmented samples:
\begin{equation}
    \text{prediction} (\td{x}_i)= \argmax_k \{ \td{\cm}(prompt(L_{k}) + \td{x}_i) \} \forall \ i \in \td{D}_{test}
\end{equation}
\section{Results \& Discussion}
We compare our approach against these baselines to show the effectiveness of our proposed entailment-style approach. 

\textbf{Majority} - Assign the majority class label to all datapoints. 

\textbf{Finetuned} - Take the already finetuned model on pre-shifted data and finetune it further on post-shift data. 

\textbf{Finetuned (post-shift only)} - Same as above, the difference being we don't finetune the model using pre-shift labels. Instead, we directly finetune using post-shift labels. This clearly indicates the benefit of entailment-style reformulation of the task where the model is able to leverage pre-shift label information alongwith post-shift label information (given the textual label descriptions/prompts) to improve over the \textit{Finetuning} approach. We don't report this for RetailQueries data since all labels are same pre-shift (irrelevant). 

\textbf{Finetuned (L1+L2)} - Finetune end-to-end with loss = cross-entropy(prediction, pre-shift label) + cross-entropy(prediction, post-shift label). This doesn't perform competitively in AGNews data which can be attributed to the fact that using both loss signals at the same time confuses the model as to what the ground-truth is. For RetailQueries, since all labels are the same (pre-shift) this boils down to same setting as previous \textit{Finetuned} approach. 

\textbf{Pre-shift Finetuning} - To demonstrate the need for models adapting to such concept shifts we add this baseline (in Figure~\ref{fig:ablation1}) which is fine-tuned only on pre-shift data and evaluated on post-shift data leading to a catastrophic drop in performance. 

\input{results}
\input{ablation1}

\input{prompts_summary}
\textbf{Our Approach (Entail-style)} - For RetailQueries, we finetune the \textit{multilingual-BERT} \cite{DBLP:conf/naacl/DevlinCLT19} model using two different entailment prompt variants based on equation \eqref{eq:prompting}. In the first variant, we create the augmented dataset by adding prompt($L_j$,k) in English as shown in Table~\ref{table2}. This leads to F1 improvements of \textbf{$\sim$1-10\%} over the finetuned baseline.
Next, in the second variant, given that the dataset is multilingual, instead of using English based prompts alone for all the input data, we add multilingual prompts\footnote{Google Translate API} corresponding to the language of the text. This leads to further gains of \textbf{$\sim$1-3\%} over the monolingual prompts. 
For AGNews shift data, we repeat the same set of experiments using \textit{multilingual-BERT}. The task-specific prompts used are mentioned in Table~\ref{table2}. We observe gains around \textbf{$\sim$3-40\%} F1 improvement using our approach. to evaluate the efficacy when labelled data is scarce and expensive, we also run in the proposed methods in the few shot settings corresponding to $N \in \{10,100,1000\}$ and observe much higher gains in low-data settings (refer Table \ref{table:results}). In general multilingual prompting has the best overall F1 with low standard deviation. This is expected as multilingual prompts makes it easier for the model to capture context when the datapoints themselves span across multiple languages. English-based prompting has the second best results.

\textbf{Effect of Random Prompts:} To tease out the contribution of informative prompts, we add random prompts by swapping \textit{E/S/C/I} with \textit{cat/lion/zebra/dog} in the prompt template mentioned above. With random prompts we observe that as we increase N the performance starts becoming comparable with it's informative prompt counterpart. This could possibly be due to the model learning some kind of mapping between the random \& actual labels with sufficient data. This phenomenon has been previously reported in \cite{explainprompt}. However, it should be noted that using random prompts resulted in a \textbf{higher standard deviation} in all the settings as evident in Figure~\ref{fig:ablation1} making such an approach unreliable for use in real-world systems.

\section{Conclusion}
\label{conclusion}
Upon deploying our final \textit{Entail-style (multilingual)} approach for adapting our internal models to the sudden concept shift, we were able to re-achieve production-level performance using just $25\%$ of pre-shift ``irrelevant" query-product pairs (now changed to E/S/C/I post-shift) thus \textbf{saving on 75\% labeling costs}. In future work, we would like to study how exactly PLMs leverage these prompts to improve their performance on downstream tasks. \cite{explainprompt} recently showed that these prompt-based learning methods don't necessarily work the way we humans think i.e. through leveraging textual semantics of the labels. Another area to explore would be starting with manual prompts and then try to learn automated prompts \cite{shin} which would eradicate the need to design prompts for varied downstream tasks further boosting the performance.

\bibliography{custom}
\bibliographystyle{acm}

\newpage
\appendix
% \section{Further Experiments}
\section{Appendix}

% \subsection{Ablations}
% \label{app:ablate}
% \input{ablation1}
% \input{prompts_summary}

\subsection{Dataset Details}
\label{app:data}
\subsubsection{RetailQueries Shift Data}
\label{app:retail}
Each of these relevance labels E/S/C/I were evaluated manually by human judges. A minimum of 3 annotations were chosen for each pair and majority vote was taken to assign the final gold label. We observe 88\% agreement rate pre-shift \& 92\% agreement post-shift across annotators on average. This is expected since the change in labeling guidelines were made to ensure it resonates better with customer intent which in turn would reduce confusion among annotators for labeling ambiguous samples. Our dataset is also multilingual spanning across 12 languages (ISO-639 codes) - English (en), Spanish (es), French (fr), Turkish (tr), Italian (it), Dutch (nl), Polish (pl), Arabic (ar), Japanese (ja), Portuguese (pt), German (de) and Hindi (hi). We release 200k query-product pairs with $\sim$100k unique queries and $\sim$180k unique products making it the first of it's kind real-world, large-scale, multi-lingual dataset to study concept shift.  
\input{retail_datastat}

Note that there might be exceptions to rules proposed in Section~\ref{datasets} to create a concept shift. For example - Brand Affinity might not matter for pharmacy related queries where showing different medicines from same brand would be irrelevant again. In future work, a detailed analysis can be done to see how well the model is able to deal with these edge-cases \& how can we further improve it.
Since only I's are changing to E/S/C/I's according to the guideline shift mentioned above, the 200k we sample are all I's according to old guidelines. It's sampled in such manner that after concept shift in new guideline 50k each will get converted to E,S,C \& I classes making it also a well-balanced dataset. Refer to Figure~\ref{fig:retail} for details.

\subsubsection{AGNews Shift Data}
\label{app:agnews}
Creating this type of shift is realistic because someone's news preferences can change with time. One might be interested in Sports news when the World Cup is going on. But later that might not be relevant. Similarly, someone might start taking interest in World news when a war is going on but otherwise wouldn't have. Hence, a model suggesting news articles to users should be able to adapt to such concept shifts to keep providing relevant suggestions without requiring huge amount of data to re-train under the new/shifted labels. Refer to Figure~\ref{fig:agnews} for details.

\input{agnews_datastat}

\subsection{Background and Related Work}
\label{related_work}

Let $\bX$ be the features (aka covariates) and $\by$ be the corresponding labels. A common assumption when deploying supervised machine learning models is that joint distribution of features and labels $\cp\mbrkt{\bX,\by}$ remains the same during the training(tr) and inference/testing(tst) stages. When this assumption is violated i.e. $\cp_{tr}\mbrkt{\bX,\by} \neq \cp_{tst}\mbrkt{\bX,\by}$, we say there is a distribution shift. Using Bayes theorem, the joint distribution can be written as product of $\cp(\bX) \cp \mbrkt{\by | \bX}$. Depending on which of these two distributions differ between training and inference stage, there are primarily two kinds of shifts. If $\cp_{tr}(\bX) \neq \cp_{tst}(\bX)$ and $\cp_{tr} \mbrkt{\by | \bX} = \cp_{tst} \mbrkt{\by | \bX}$, then there is covariate shift. If $\cp_{tr}(\bX) = \cp_{tst}(\bX)$ and $\cp_{tr} \mbrkt{\by | \bX} \neq \cp_{tst} \mbrkt{\by | \bX}$, then there is concept shift. Often times, it possible that both these shifts can occur together. Concept shift is relatively understudied as compared to covariate shift, nevertheless it is still an important problem to consider while deploying machine learning classifiers \cite{xu2021concept}. We refer to \cite{moreno2012unifying} for a comprehensive introduction to the different data shifts that can occur in real-world applications. In this work, we focus on tackling concept shift. Also, based on the speed of change, concept shift can manifest itself in two major forms - \textit{gradual shift} and \textit{sudden shift}\cite{lindstrom2010handling,narasimhamurthy2007framework,suarez2022survey}.
% Sudden Shift occurs when there are abrupt changes in the concept. For example, in a news filtering usecase, the break-out of a deadly pandemic can make articles about vaccines/biotech advancements more relevant for the user, which were previously irrelevant.
% Gradual Shift, as the name refers to, is a more slow, continuous and evolved changes in the concept. An example of Gradual Shift is how meanings of words evolve over time. For example, the meanings of the words `BERT' and `ELMO' have changed from persons to neural network architectures in the recent past \cite{bjerva2020back}. 
Although there are other kinds of shifts like incremental shifts, recurring shifts, etc. in this work, \textbf{we restrict ourselves to the sudden shift case} in the context of text classification as this is relatively more challenging because the timeline to acquire labelled data is more stringent.

One of the straightforward ways to tackle concept shift is to retrain the model from scratch using the new data, as and when there is a concept shift. If there is no explicit information that a concept shift has occurred, then monitoring using a concept drift detector may be required to decide when to retrain the model \cite{lu2018learning}. However, training PLMs requires that there is enough labelled data \cite{vaswani2017attention, DBLP:conf/naacl/DevlinCLT19,raffel,gpt,thoppilan,rae,chowdhery} which are expensive and time consuming to obtain. Previous approaches to tackle sudden changes in concept have relied on giving importance to datapoints based on their recency, either through a weightage/discounting factor or only considering a recent window of data whilst discarding everything prior to that window \cite{klinkenberg2002concept, delany2005case}. A longer window is more suitable for the gradual shift case, where as a shorter window is more effective for fast changing environments. The time window can be fixed or adaptive \cite{widmer1993effective}. In the context of PLMs, fine-tuning based on a new batch of data can considered as a sliding-windowing approach. We use this as one of the baselines in our paper. Also as mentioned earlier we limit the scope of this work to only sudden shift (since it's more challenging), we don't compare against other windowing-based methods as those are suited for gradual shift problems where there is some timeline-based information associated with the shifts. For a more detailed review, we refer to the study by \cite{lu2018learning}.

\subsection{Implementation Details}
\label{sec:impl}
We use the Multilingual BERT-base\footnote{\url{https://huggingface.co/bert-base-multilingual-cased}} model as our backbone model for all experiments. It has total of 110M trainable parameters, 12 heads, 12 layers and 768 hidden dimension. On top of that we have a single linear layer for final classification. We use the AdamW optimizer with linear learning rate decay post 10\% warmup steps for finetuning. We use the standard cross-entropy loss for finetuning purposes. 

For RetailQueries dataset, we train all models for 5 epochs using learning rate 1e-5, max sequence length 128 \& batch size 16. It is a two-sentence task since we have both query \& product information to be classified into E/S/C/I. The input to the model (also definition of \textit{concat()} operation) is in the format - $query$ [SEP] $prompt(label)$ [SEP] $product$. As mentioned in Section~\ref{proposed_approach}, we have 1 positive example and (K-1) negative examples per datapoint. So to reduce the label imbalance, we augment one more positive sample created by random deletion of 5\% text span in product title. For AGNews dataset, we train for 10 epochs using learning rate 1e-6, max sequence length 128 \& batch size 16. It's a single-sentence task of classifying news articles into relevant/irrelevant news. So the input format is - $prompt(label)$ [SEP] $news$. Here choosing the position where prompt is appended is a design choice. We observe best results in the above reported settings. 

We ran all experiments for 5 different seeds on 4 Nvidia V100 GPUs parallely and report the mean and standard deviations. Training takes around $\sim$5 hrs on RetailQueries \& $\sim$1 hr on AGNews datasets to reproduce the reported results.
%%%%%%%%%%%%%%%%%%%%%%%%%%%%%%%%%%%%%%%%%%%%%%%%%%%%%%%%%%%%

\subsection{Limitations}
While prompting-based methods outperform direct supervised baselines by a large margin it comes at an expense of increased inference time. However, given the prohibitively large cost of getting reliable human annotations for new data, and the significant gains observed in low-data settings (see N=1000 in Table~\ref{table:results}) it seems like a sensible trade-off. Moreover with carefully designed prompts, even smaller PLMs could potentially achieve similar performance with lesser data \cite{goodprompt} thus reducing inference time. Also, reducing the  model parameter precision could further speed up inference (maybe with some performance trade-off) on specialized hardware. We leave these discussions to be explored in future work.

\subsection{Ethical Considerations}
We report only aggregated results in the main paper. We have not or do not intend to share any Personally Identifiable Information (PII) in our released dataset or in the paper. We use standard Huggingface\footnote{https://huggingface.co/} libraries for training our models to promote reproducible research. But it must be kept in mind that these models are not to be used for generation purposes (like GPT \cite{gpt}). Using biased prompts might lead the model to generate biased responses given these large language models are pre-trained using publicly available data which is why we do not intend to release the trained model weights.

\end{document}

%% file: shortcuts.tex
\def\by{\mathbf{y}}

\def\bX{\mathbf{X}}

\def\b0{\mathbf{0}}
\def\b1{\mathbf{1}}

\DeclareMathOperator*{\argmax}{arg\,max}

\newcommand{\mbrkt}[1]{\left(#1\right)}
\newcommand{\sqrbrkt}[1]{\{#1\}}
   \newcommand{\cd}{\mathcal{D}}        \newcommand{\cl}{\mathcal{L}} \newcommand{\cm}{\mathcal{M}}   \newcommand{\cp}{\mathcal{P}}         

\newcommand{\td}[1]{\tilde{#1}}

%% file: anecdotes.tex
\begin{table*}[h]
\centering
\small
% \begin{tabular}{|c|c|c|c|p{1cm}p{1cm}p{1cm}p{1cm}p{1cm}p{1cm}p{1cm}|}
\begin{tabular}{|p{6.5cm}@{\hskip 0.05in}|p{6.5cm}|}
\hline
% \toprule
% \textbf{}   & \textbf{Label Shift} & \textbf{Prompt (Label)}   \\ 
% \hline 
\multicolumn{2}{|c|}{\textbf{RetailQueries Shift Data (Real)}}\\
\hline
 \hfil \textbf{Query/Product}  &  \hfil \textbf{Pre-shift $\rightarrow$ Post-shift}     \\ 
\hline
 vacation things / Bug Bite Suction Tool & Irrelevant $\rightarrow$ Exact (utility same) \\
 peppa pig muddy puddle / plush toy peppa pig & Irrelevant $\rightarrow$ Substitute (brand same) \\
 dove soap / dove hand wash nourishing, 3x500ml & Irrelevant $\rightarrow$ Complement (brand same + utility related) \\
 small globe / Clamp desk lamp & Irrelevant $\rightarrow$ Irrelevant (no match) \\

\hline
\multicolumn{2}{|c|}{\textbf{AGNews Shift Data (Synthetic)}}\\
\hline
  \hfil \textbf{Title}  &  \hfil \textbf{Pre-shift $\rightarrow$ Post-shift}     \\ 
\hline
 Australia march into dominant position against NZ & Irrelevant $\rightarrow$ Relevant (Sports) \\

 Timing Of Indian Move In Kashmir Vital: Pak Paper & Relevant $\rightarrow$ Irrelevant (World)\\
 Stocks to Watch Tuesday & Relevant $\rightarrow$ Relevant (Business)\\
\hline
\end{tabular}
\caption{\small \textbf{A Few anecdotal examples} from the datasets we use for our study. For \textit{(top)} RetailQueries, we provide the reasons for shift based on labeling guideline change. For \textit{(bottom)} AGNews, we provide the topics under which shift happens.}
\label{table1}
\end{table*}

%% file: main_figure.tex
\begin{figure*}[t]
\centering
\centerline{\includegraphics[width=0.9\linewidth]{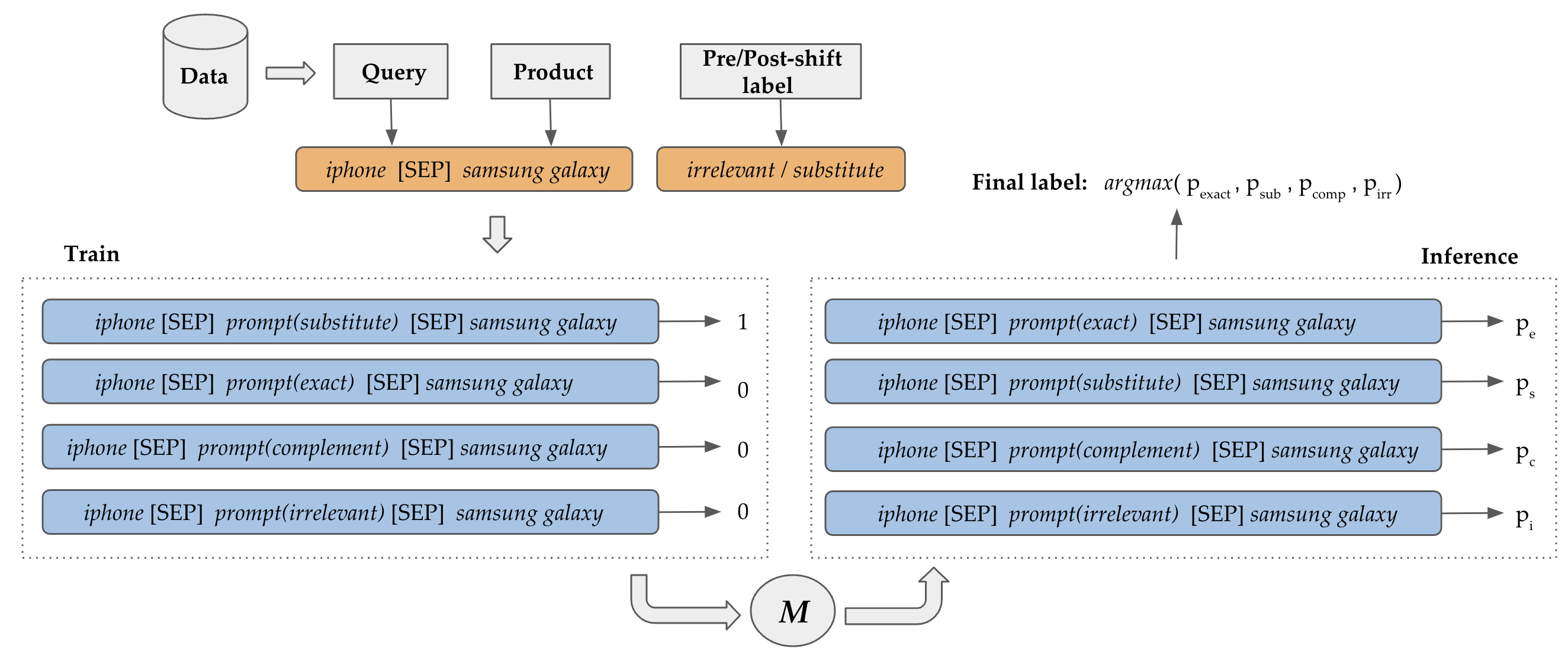}}
\caption{
\small
 [Best viewed in color] \textbf{Proposed Model}. Consider the (query, product) pair \textit{(iphone, samsung galaxy)} to be earlier ``irrelevant'' and now ``substitute" in the concept-shifted training data. Then we create 3 -ve examples \textit{(iphone +  samsung galaxy, changed to exact, 0)}, \textit{(iphone + samsung galaxy, changed to complement, 0)} , \textit{(iphone + samsung galaxy, remained irrelevant, 0)} and 1 +ve example \textit{(iphone + samsung galaxy, changed to substitute, 1)}. Once the model $\cm$  is fine-tuned on the concept-shifted training data, during inference, say we encounter the  (query, product) pair \textit{(earphones, airpods pro)}. We compute the argmax of probabilities of the following augmented inputs: \textit{(earphones + airpods pro, exact/substitute/complement/irrelevant)} and pick the predicted label accordingly.}
\label{fig:illustration}
\end{figure*}

%% file: results.tex
\begin{table*}[ht]
\centering
\small
% \begin{tabular}{p{3.25cm}p{2.25cm}p{2.25cm}p{2.25cm}p{2.25cm}} 
\begin{tabular}{c@{\hskip 0.1in}c@{\hskip 0.1in}c@{\hskip 0.1in}c@{\hskip 0.1in}c} 
\hline
\textbf{Model  }   & \multicolumn{4}{c}{\textbf{Macro Avg. F1-score  }}    \\\cmidrule{2-5} 
% [0.5ex]
% \hline \\
% & & & & \\ \cmidrule{2-5}
& Full Data & N = 1000 & N = 100 & N = 10 \\\cmidrule{2-5}

\hline 
\multicolumn{5}{c}{\textbf{RetailQueries Shift Data (Real)}}\\
\hline
% \multicolumn{3}{c}{\textbf{\latenthatred}}\\
% \hline 
Majority & 10.0(0.0) & 10.0(0.0) & 10.0(0.0) & 10.0(0.0) \\
Finetuned & 51.27(0.17) & 25.67(0.56) & 19.58(0.90) & 18.53(1.73) \\
Finetuned (L1+L2) & 51.25(0.11) & 25.7(0.6) & 18.45(0.88) & 18.12(1.5) \\
% \hline
Entail-style (english) & \underline{51.67}(0.08) & \underline{29.62}(1.10) & \underline{24.37}(1.28) & \underline{23.83}(0.37) \\
% \hline
\textbf{Entail-style (multilingual)} & \textbf{52.62(0.33)} & \textbf{32.32(0.37)} & \textbf{26.36(0.34)} & \textbf{24.93(0.44)} \\
% \hline
% Prompting (random) & 51.20(0.90) & 27.07(2.13) & 21.86(1.49) & 20.87(1.1) \\
\hline

\multicolumn{5}{c}{\textbf{AGNews Shift Data (Synthetic)}}\\
\hline 
% & Full Data & k = 1000 & k = 100 & k = 10 \\
% \hline
% \multicolumn{3}{c}{\textbf{\latenthatred}}\\
% \hline 
Majority & 33.33(0.0) & 33.33(0.0) & 33.33(0.0) & \underline{33.33}(0.0) \\
% Not Finetuned  & 22.89(0.59) & 22.89(0.59) & 22.89(0.59) & 22.89(0.59) \\
Finetuned & \underline{93.47}(0.09) & \underline{88.31}(0.78) & \underline{42.7}(0.09) & 24.35(0.80) \\
Finetuned (post-shift only) & 91.55(0.23) & 86.27(0.5) & 38.6(0.66) & 18.19(0.89)\\
% \hline
Finetuned (L1+L2) & 56.97(0.92) & 54.73(2.95) & 38.43(0.98) & 24.42(0.93) \\
% \hline
\textbf{Entail-style (english)} & \textbf{96.48(0.13)} & \textbf{93.5(0.10)} & \textbf{83.61(0.91)} & \textbf{73.14(3.22)} \\
% \hline
% Prompting (random) & 96.42(0.09) & 93.09(0.50) & 74.86(4.85) & 61.84(5.61) \\
\hline

% \hline
% \bertaug & Mul & 63.8 & 58.6 \\
% \hline
% \multicolumn{3}{c}{\textbf{AGNews dataset}}\\
% \hline 
% Baseline & 69.00 & 67.40 \\ %68.3, 66.98 %& 67.83 & 66.69
% % \hline
% +\ssmixup & 69.59 & 67.72 \\
% % \hline
% +\easymix & \textbf{69.70} & \textbf{68.66} \\
% \hline
\end{tabular}
\caption{
\small
\textbf{Results on RetailQueries \& AGNews Shift datasets} for various few-shot settings (N=10, 100, 1000, full data). Best runs are marked in \textbf{bold} and second best is \underline{underlined}. We report the average and standard deviation of 5 runs (reported results are significantly different: M-W-U test with p-value < 0.05).}
\label{table:results}
\end{table*}

%% file: ablation1.tex
\begin{figure*}[h]

  \centering
  \centerline{\includegraphics[width=0.8\linewidth]{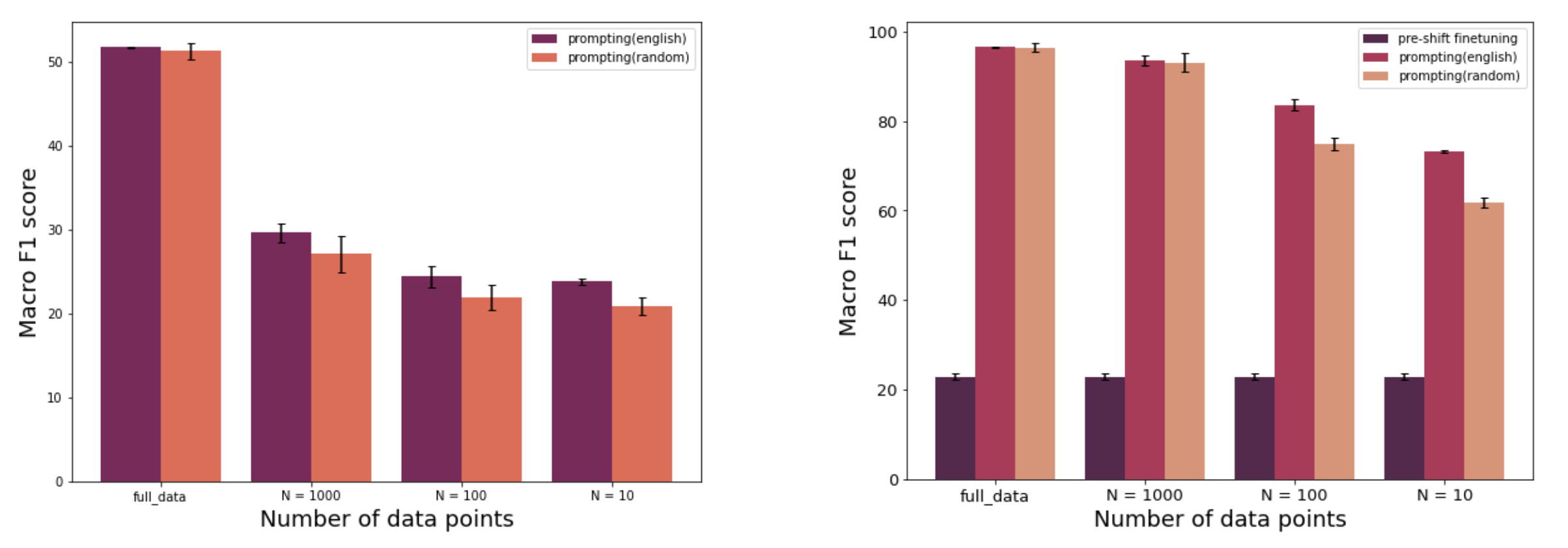}}
  % \centerline{\small Dataset distribution}
\caption{\small [Best viewed in color] \textbf{Random vs Informative Prompts.} We also add results for Pre-Shift finetuning to show the drastic drop in performance compared to the entailment-style approaches (Note: Pre-Shift Finetuning for RetailQueries is not shown since all labels were irrelevant before). The macro-F1 scores reported are average of 5 runs. The bars indicate standard deviation (reported results are significantly different: M-W-U test with p-value < 0.05). \textit{(Left)} RetailQueries Dataset. \textit{(Right)} AGNews Dataset.}
%\am{The reviewer asked to release the data corresponding to fig (d). They wanted the actual names for these groups and the data annotated by under each group.}
\label{fig:ablation1}
\end{figure*}

%% file: prompts_summary.tex
\begin{table*}[htbp]
\centering
\small
% \begin{tabular}{|c|c|c|c|p{1cm}p{1cm}p{1cm}p{1cm}p{1cm}p{1cm}p{1cm}|}
\begin{tabular}{c@{\hskip 0.1in}l@{\hskip 0.05in}l}
% \hline
\toprule
\textbf{Language}   & \textbf{Label Shift} & \textbf{Prompt (Label)}   \\ 
\hline 
\multicolumn{3}{c}{\textbf{RetailQueries Shift Data (Real)}}\\
\hline
\multirow{ 4}{*}{English} & Exact & changed to exact match \\
& Substitute & changed to substitute match \\
& Complement & changed to complement match \\
& Irrelevant & remained irrelevant match \\
\hline
\multirow{4}{*}{Spanish} & Exact & cambiado a coincidencia exacta \\
& Substitute & cambiado para sustituir el partido \\
& Complement & cambiado para complementar la coincidencia \\
& Irrelevant & permaneció un partido irrelevante  \\
\hline
\multicolumn{3}{c}{\textbf{AGNews Shift Data (Synthetic)}}\\ \hline
\multirow{ 4}{*}{English} & Irrelevant $\rightarrow$ Relevant & changed to relevant news \\
& Irrelevant $\rightarrow$ Irrelevant & remained irrrelvant news \\
& Relevant $\rightarrow$ Irrelevant & changed to irrelevant news \\
& Relevant $\rightarrow$ Relevant & remained relevant news \\
\hline
\end{tabular}
\caption{\small For RetailQueries shift dataset (multilingual), we mention only the post-shift labels since pre-shift label is `irrelevant' for all. We run both plain English-based and Multilingual prompting. The prompts for English and Spanish languages are given above; similar prompts for other languages have been obtained using Google Translate. For AGNews shift dataset (English), the prompts contain information about both pre \& post-shift labels. }
\label{table2}
\end{table*}

%% file: retail_datastat.tex
\begin{figure}[!h]

  \centering
  \centerline{\includegraphics[width=\linewidth]{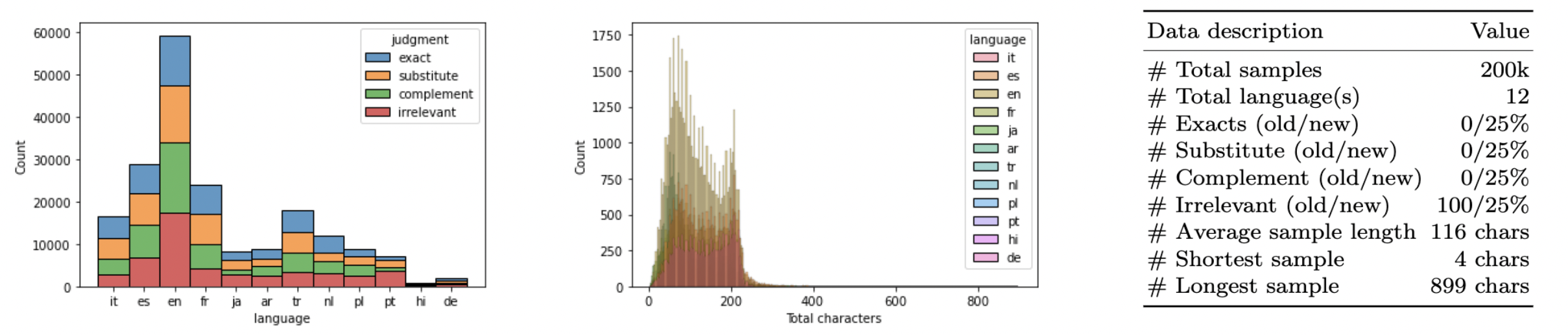}}
  % \centerline{\small Dataset distribution}
\caption{\small [Best viewed in color] RetailQueries Data \textit{(Left)} Post-Shift label distribution (pre-shift all were irrelevants -Section~\ref{datasets}) \textit{(Center)} Character length distribution. \textit{(Right)} Dataset Statistics.}
%\am{The reviewer asked to release the data corresponding to fig (d). They wanted the actual names for these groups and the data annotated by under each group.}
\label{fig:retail}
\end{figure}

%% file: agnews_datastat.tex
\begin{figure}[h]

  \centering
  \centerline{\includegraphics[width=\linewidth]{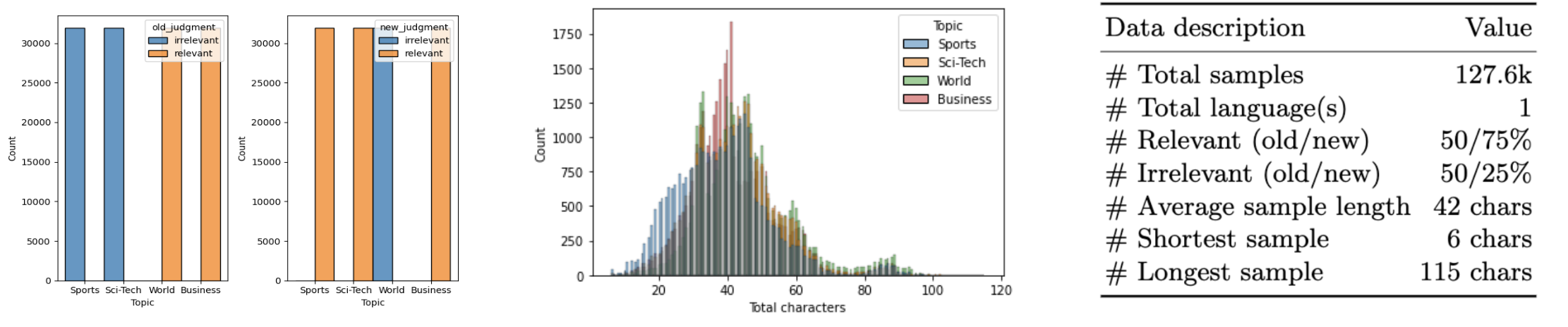}}
  % \centerline{\small Dataset distribution}
\caption{\small [Best viewed in color] AGNews Data \textit{(Left)} Pre \& Post-Shift label distribution \textit{(Center)} Character length distribution. \textit{(Right)} Dataset Statistics.}
%\am{The reviewer asked to release the data corresponding to fig (d). They wanted the actual names for these groups and the data annotated by under each group.}
\label{fig:agnews}
\end{figure}

%% file: neurips_distshift_2023.bbl
\begin{thebibliography}{10}

\bibitem{bjerva2020back}
{\sc Bjerva, J., Kouw, W., and Augenstein, I.}
\newblock Back to the future--temporal adaptation of text representations.
\newblock In {\em Proceedings of the AAAI Conference on Artificial
  Intelligence\/} (2020), vol.~34, pp.~7440--7447.

\bibitem{gpt}
{\sc Brown, T.~B., Mann, B., Ryder, N., Subbiah, M., Kaplan, J., Dhariwal, P.,
  Neelakantan, A., Shyam, P., Sastry, G., Askell, A., Agarwal, S.,
  Herbert-Voss, A., Krueger, G., Henighan, T., Child, R., Ramesh, A., Ziegler,
  D.~M., Wu, J., Winter, C., Hesse, C., Chen, M., Sigler, E., Litwin, M., Gray,
  S., Chess, B., Clark, J., Berner, C., McCandlish, S., Radford, A., Sutskever,
  I., and Amodei, D.}
\newblock Language models are few-shot learners, 2020.

\bibitem{chowdhery}
{\sc Chowdhery, A., Narang, S., Devlin, J., Bosma, M., Mishra, G., Roberts, A.,
  Barham, P., Chung, H.~W., Sutton, C., Gehrmann, S., et~al.}
\newblock Palm: Scaling language modeling with pathways.
\newblock {\em arXiv preprint arXiv:2204.02311\/} (2022).

\bibitem{delany2005case}
{\sc Delany, S.~J., Cunningham, P., Tsymbal, A., and Coyle, L.}
\newblock A case-based technique for tracking concept drift in spam filtering.
\newblock In {\em Applications and Innovations in Intelligent Systems XII:
  Proceedings of AI-2004, the Twenty-fourth SGAI International Conference on
  Innovative Techniques and Applications of Artificial Intelligence\/} (2005),
  Springer, pp.~3--16.

\bibitem{DBLP:conf/naacl/DevlinCLT19}
{\sc Devlin, J., Chang, M., Lee, K., and Toutanova, K.}
\newblock {BERT:} pre-training of deep bidirectional transformers for language
  understanding.
\newblock In {\em Proceedings of the 2019 Conference of the North American
  Chapter of the Association for Computational Linguistics: Human Language
  Technologies, {NAACL-HLT} 2019, Minneapolis, MN, USA, June 2-7, 2019, Volume
  1 (Long and Short Papers)\/} (2019), J.~Burstein, C.~Doran, and T.~Solorio,
  Eds., Association for Computational Linguistics, pp.~4171--4186.

\bibitem{goodprompt}
{\sc Jin, W., Cheng, Y., Shen, Y., Chen, W., and Ren, X.}
\newblock A good prompt is worth millions of parameters: Low-resource
  prompt-based learning for vision-language models.
\newblock In {\em Proceedings of the 60th Annual Meeting of the Association for
  Computational Linguistics (Volume 1: Long Papers)\/} (Dublin, Ireland, May
  2022), Association for Computational Linguistics, pp.~2763--2775.

\bibitem{klinkenberg2004learning}
{\sc Klinkenberg, R.}
\newblock Learning drifting concepts: Example selection vs. example weighting.
\newblock {\em Intelligent data analysis 8}, 3 (2004), 281--300.

\bibitem{klinkenberg2002concept}
{\sc Klinkenberg, R., and R{\"u}ping, S.}
\newblock Concept drift and the importance of examples.
\newblock In {\em Text mining--theoretical aspects and applications\/} (2002),
  Citeseer.

\bibitem{kolter2007dynamic}
{\sc Kolter, J.~Z., and Maloof, M.~A.}
\newblock Dynamic weighted majority: An ensemble method for drifting concepts.
\newblock {\em The Journal of Machine Learning Research 8\/} (2007),
  2755--2790.

\bibitem{lindstrom2010handling}
{\sc Lindstrom, P., Delany, S.~J., and Mac~Namee, B.}
\newblock Handling concept drift in a text data stream constrained by high
  labelling cost.
\newblock In {\em Twenty-Third International FLAIRS Conference\/} (2010).

\bibitem{lu2018learning}
{\sc Lu, J., Liu, A., Dong, F., Gu, F., Gama, J., and Zhang, G.}
\newblock Learning under concept drift: A review.
\newblock {\em IEEE transactions on knowledge and data engineering 31}, 12
  (2018), 2346--2363.

\bibitem{moreno2012unifying}
{\sc Moreno-Torres, J.~G., Raeder, T., Alaiz-Rodr{\'\i}guez, R., Chawla, N.~V.,
  and Herrera, F.}
\newblock A unifying view on dataset shift in classification.
\newblock {\em Pattern recognition 45}, 1 (2012), 521--530.

\bibitem{muller2020addressing}
{\sc M{\"u}ller, M., and Salath{\'e}, M.}
\newblock Addressing machine learning concept drift reveals declining vaccine
  sentiment during the covid-19 pandemic.
\newblock {\em arXiv preprint arXiv:2012.02197\/} (2020).

\bibitem{narasimhamurthy2007framework}
{\sc Narasimhamurthy, A.~M., and Kuncheva, L.~I.}
\newblock A framework for generating data to simulate changing environments.
\newblock In {\em Artificial intelligence and applications\/} (2007),
  pp.~415--420.

\bibitem{rae}
{\sc Rae, J.~W., Borgeaud, S., Cai, T., Millican, K., Hoffmann, J., Song, F.,
  Aslanides, J., Henderson, S., Ring, R., Young, S., et~al.}
\newblock Scaling language models: Methods, analysis \& insights from training
  gopher.
\newblock {\em arXiv preprint arXiv:2112.11446\/} (2021).

\bibitem{raffel}
{\sc Raffel, C., Shazeer, N., Roberts, A., Lee, K., Narang, S., Matena, M.,
  Zhou, Y., Li, W., and Liu, P.~J.}
\newblock Exploring the limits of transfer learning with a unified text-to-text
  transformer, 2020.

\bibitem{roy2023hate}
{\sc Roychowdhury, S., and Gupta, V.}
\newblock Data-efficient methods for improving hate speech detection.
\newblock In {\em Findings of the Association for Computational Linguistics:
  EACL 2023\/} (2023), pp.~125--132.

\bibitem{shin}
{\sc Shin, T., Razeghi, Y., Logan~IV, R.~L., Wallace, E., and Singh, S.}
\newblock {A}uto{P}rompt: {E}liciting {K}nowledge from {L}anguage {M}odels with
  {A}utomatically {G}enerated {P}rompts.
\newblock In {\em Proceedings of the 2020 Conference on Empirical Methods in
  Natural Language Processing (EMNLP)\/} (Online, Nov. 2020), Association for
  Computational Linguistics, pp.~4222--4235.

\bibitem{suarez2022survey}
{\sc Su{\'a}rez-Cetrulo, A.~L., Quintana, D., and Cervantes, A.}
\newblock A survey on machine learning for recurring concept drifting data
  streams.
\newblock {\em Expert Systems with Applications\/} (2022), 118934.

\bibitem{thoppilan}
{\sc Thoppilan, R., De~Freitas, D., Hall, J., Shazeer, N., Kulshreshtha, A.,
  Cheng, H.-T., Jin, A., Bos, T., Baker, L., Du, Y., et~al.}
\newblock Lamda: Language models for dialog applications.
\newblock {\em arXiv preprint arXiv:2201.08239\/} (2022).

\bibitem{vaswani2017attention}
{\sc Vaswani, A., Shazeer, N., Parmar, N., Uszkoreit, J., Jones, L., Gomez,
  A.~N., Kaiser, {\L}., and Polosukhin, I.}
\newblock Attention is all you need.
\newblock {\em Advances in neural information processing systems 30\/} (2017).

\bibitem{explainprompt}
{\sc Webson, A., and Pavlick, E.}
\newblock Do prompt-based models really understand the meaning of their
  prompts?, 2022.

\bibitem{widmer1993effective}
{\sc Widmer, G., and Kubat, M.}
\newblock Effective learning in dynamic environments by explicit context
  tracking.
\newblock {\em Lecture Notes in Computer Science\/} (1993), 227--227.

\bibitem{xu2021concept}
{\sc Xu, Y., and Klabjan, D.}
\newblock Concept drift and covariate shift detection ensemble with lagged
  labels.
\newblock In {\em 2021 IEEE International Conference on Big Data (Big Data)\/}
  (2021), IEEE, pp.~1504--1513.

\bibitem{agnews}
{\sc Zhang, X., Zhao, J., and LeCun, Y.}
\newblock Character-level convolutional networks for text classification, 2015.

\end{thebibliography}
